\documentclass[11pt]{article}
\usepackage{coling2016}
\usepackage{times}
\usepackage{url}
\usepackage{latexsym}
\usepackage{multirow}
\usepackage{epstopdf}
\usepackage{amssymb}
\usepackage{booktabs}
\usepackage{multirow}
\usepackage{amsmath}

\title{Bidirectional LSTM-CRF for Clinical Concept Extraction}

\author{Raghavendra  Chalapathy \\ University of Sydney\\Capital Markets CRC\\
	Darlington NSW 2008 \\ rcha9612@uni.sydney.edu.au
	\And  
	Ehsan Zare Borzeshi  \\ Capital Markets CRC\\
	3/55 Harrington St \\Sydney NSW 2000 \\ ezborzeshi@cmcrc.com
	\And
	Massimo Piccardi \\University of Technology Sydney \\ PO Box 123 \\  Broadway NSW 2007 \\
	Massimo.Piccardi@uts.edu.au
}

\begin{document}
\maketitle
\begin{abstract}

Extraction of concepts present in patient clinical records is an essential step in clinical research. The 2010 i2b2/VA Workshop on Natural Language Processing Challenges for clinical records presented concept extraction (CE) task,  with aim to identify concepts (such as treatments, tests, problems) and classify them into predefined categories. State-of-the-art CE approaches heavily rely on hand crafted features and domain specific resources which are hard to collect and tune. For this reason, this paper employs bidirectional LSTM with CRF decoding   initialized with general purpose off-the-shelf word embeddings for  CE. The experimental results achieved on 2010 i2b2/VA reference standard corpora using bidirectional LSTM CRF ranks closely with  top ranked systems.

\end{abstract}

\section{Introduction}
\label{intro}

Patient clinical records contain  longitudinal record of patient health, disease, test's conducted  and response to treatment, often useful for epidemiologic and clinical research. Thus extracting these information  has been of immense value for both clinical practise and to improve  quality of patient care provided  while reducing healthcare costs. Concept extraction (CE) aims to identify medical concept mentions such as problems, test, treatments in  clinical  records (Eg: discharge summaries, progress reports) and classify them into pre-defined categories. The concepts in  clinical records are often expressed with unstructured free text, rendering their extraction a daunting task for  clinical Natural Language Processing (NLP) systems. The CE  problem is analogous to well-studied Named Entity Recognition (NER) task in general NLP domain. Traditional approaches to extracting concepts relied on  rule based systems or dictionaries (lexicon's) using string comparision to recognise concepts of interest. The concepts  represent drug names, anatomical nomenclature, other specialised names and phrases which are not part of mundane English vocabulary. For instance "resp status" should be interpreted as "response status". Furthermore the use of abbreviated phrases are very common among medical fraternity and many of these abbreviations have alternative meanings in other genres of English. Intrinsically, rule based systems are hard to scale, and ineffective in the presence of informal sentences and abbreviated phrases~\cite{liu2015drug}. Dictionary based systems perform a  fast look-up  from medical ontologies such as Unified Medical Language System (UMLS)  to extract concepts~\cite{kipper2008system}. Although these systems achieve high precison but suffer from low recall ( i,e they may not identify significant number of concepts) due to missplelled words or medical jargons not present in dictionaries. To overcome these limitations various supervised and semi-supervised machine learning (ML) approaches and its variants  have been proposed utilizing conditional random fields (CRF), maximum entropy and support vector machines~(SVM) models which utilize both textual and contextual information while reducing the dependency on lexicon lookup~\cite{lafferty2001conditional,berger1996maximum,joachims1998text}. However these state-of-the-art ML approaches follow two step process of domain specific feature engineering and classification, which are highly dedicated hand-crafted systems and require labour intensive expert knowledge. For this reason, this paper employs bidirectional LSTM-CRF intialized with  general purpose off-the-shelf neural word embeddings derived from Glove~\cite{Pennington:14} and Word2Vec~\cite{Mikolov:13}  for automatic feature learning thus avoiding time-consuming feature engineering,  which deliver system performance comparable to the best submissions from the 2010 i2b2/VA challenge.

\section{Related Work}
\label{relatedworks}

Most of the research to date have formulated CE as a sequence labelling NER problem employing various supervised and semi-supervised ML algorithms employing focussed domain-dependent attributes and specialized text features~\cite{uzuner20112010}. Similarly hybrid models obtained by cascading CRF and SVM algorithms along with several pattern matching rules are shown to produce effective results~\cite{boagcliner}. The efficacy of including pre-processing technique (such as truecasing and annotation combination) along with CRF based NER system to improve concept extraction performace was exemplified by \cite{fu2014improving}. The best performing system for 2010 i2b2/VA concept extract task adopted unsupervised feature representations  derived from unlabeled corpora using Brown clustering technique along with semi-supervised Markov HMM models~\cite{de2011machine}. However, the unsupervised one-hot word feature representations derived from Brown clustering fails to capture multiple aspect relation between words. Subsequently~\cite{jonnalagadda2012enhancing} demonstrated that random indexing model with distributional word representations  improve clinical concept extraction. With recent success of incorporating word embeddings derived from the entire English wikipedia in various NER task \cite{collobert2011natural}, binarized word embeddings derived from domain specific copora (Eg: Monitoring in Intensive Care (MIMIC) II corpus)  has improved performance of CRF based concept extraction system~\cite{wu2015study}. In the broader field of machine learning, the recent years have witnessed proliferation of  deep neural networks, with unprecedented results in tasks such as visual, speech and NER. One of the main advantages of neural networks is that they learn features automatically thus avoiding laborious feature engineerin. Given these promising results obtained the main goal of this paper is to employ bidirectional LSTM CRF intialized with general off-the-shelf unsupervised word embeddings derived from Glove and Word2Vec models and evaluate its performance. The experimental results obtained on 2010 i2b2/VA reference standard corpora without use of any extensive feature engineering  and domain specific resources is very encouraging.

%
%
\blfootnote{
    %
    %
    \hspace{-0.65cm}  
   
    %

}

\begin{table*}[ht]
	\small
	\centering
	
	\begin{tabular}{|c|c|c|c|c|c|c|c|c|c|c|c|}
		\hline \bf Sentence & \textit{His}& \textit{HCT} & \textit{had}& \textit{dropped} & \textit{from} &\textit{36.7} &\textit{despite} &\textit{2U} &\textit{PRBC} &\textit{and} &\textit{3U-FFP} \\ 
		\hline \textbf{Concept class}& \textit{B-test}& \textit{I-test}& \textit{O} & \textit{O}& \textit{O} & \textit{O} & \textit{ O} & \textit{B-treatment} & \textit{I-treatment} & \textit{O} & \textit{O}\\ 			
		\hline
	\end{tabular}
	\caption{Example sentence in a CE task with concept classes represented in IOB format.}
	\label{table1} 
\end{table*}

\begin{table*}[ht]	
	\centering

		\begin{tabular}{|c|c|c|}
			\hline
			\multirow{3}{*}{} & 
			\multicolumn{2}{c|}{\bf {\small 2010 i2b2/VA}} \\  
			\cline{2-3}
		
			& Training for CE task & Test for CE \\		
			\hline
			notes &$170$ &$256$  \\
			sentences &$16315$&$27626$\\ 
			\hline		
			problem  & $7073$   & $12592$  \\		
			test&  $4608$ & $9225$\\		
			treatment& $4844$ & $9344$\\
			 \hline
		\end{tabular}
		\caption{Statistics of training and test datasets used for 2010-i2b2 concept extraction.}
		\label{table2} 
	\end{table*}

\section{ The Proposed Approach}
CE can be formulated as a joint segmentation and classification task over a predefined set of classes. As an example, consider the input sentence provided in Table~\ref{table1}. The notation follows the widely adopted in/out/begin (IOB) entity representation with, in this instance, \textit{HCT} as the test, \textit{2U PRBC} as the treatment. In this paper, we approach the CE task by bidirectional LSTM CRF and we therefore provide a brief description hereafter. In a bidirectional LSTM CRF, each word in the input sentence is first mapped to a random real-valued  vector of arbitrary dimension, $d$. Then, a measurement for the word, noted as $x(t)$, is formed by concatenating the word's own vector with a window of preceding and following vectors (the ``context''). An example of input vector with a context window of size $s = 3$ is:
\begin{equation}
	\begin{split}
		w_{3}(t) = [His, \textbf{HCT}, dropped], \\
		`His' \rightarrow x_{HCT} \in \mathbb{R}^{d}, \\
		`HCT' \rightarrow x_{His} \in \mathbb{R}^{d}, \\
		`dropped' \rightarrow x_{dropped} \in \mathbb{R}^{d}, \\
		x(t) = [x_{His}, x_{\textbf{HCT}}, x_{dropped}] \in \mathbb{R}^{3d}
	\end{split}
\end{equation}

\noindent where $w_{3}(t)$ is the context window centered around the $t$-th word, $'HCT'$, and $x_{word}$ represents the numerical vector for $word$.

\subsection{Word Embeddings}
Word embeddings are  dense vector  representations of natural language words that preserves the semantic and syntactic similarities between them. The vector representations could be generated by either count based such as Hellinger-PCA ~\cite{lebret2013word}, direct prediction models such as Word2Vec comprising of Skip-gram or Common Bag of Words (CBOW) or Glove word embeddings. Glove vector representations captures complex patterns beyond word similarity through by combining  efficient use of word co-occurance statistics and generate a global vector representation for any given word.


\subsection{Bidirectional LSTM-CRF Networks}

The LSTM was designed to overcome this limitation by incorporating a gated memory-cell to capture long-range dependencies within the data~\cite{hochreiter1997long}. In the bidirectional LSTM, for any given sentence, the network computes both a left, $\overrightarrow{h}(t)$, and a right, $\overleftarrow{ h}(t)$, representations of the sentence context at every input, $x(t)$. The final representation is created by concatenating them as $h(t) = [\overrightarrow{h}(t)$;$\overleftarrow{ h}(t)]$. All these networks utilize the $h(t)$ layer as an implicit feature for entity class prediction: although this model has proved effective in many cases, it is not able to provide joint decoding of the outputs in a Viterbi-style manner (e.g., an I-group cannot follow a B-brand; etc). Thus, another modification to the bidirectional LSTM is the addition of a conditional random field (CRF)~\cite{lafferty2001conditional} as the output layer to provide optimal sequential decoding. The resulting network is commonly referred to as the bidirectional LSTM-CRF \cite{lample2016neural}.

\begin{table*}[ht]	
	\centering

		\begin{tabular}{|c|c|c|c|}
			\hline
			\multirow{3}{*}{Methods} & 
			\multicolumn{3}{c|}{\bf {\small 2010 i2b2/VA}} \\
			\cline{2-4}
			& Precision & Recall & F$_1$ Score \\
			\hline
			semi-supervised Markov HMM \cite{de2011machine} &$86.88$ &$83.64$ & $85.23$ \\
			distributonal semantics-CRF \cite{jonnalagadda2012enhancing} &$85.60$&$82.00$ &$83.70$  \\	
			binarized neural embedding CRF\cite{wu2015study}&$85.10$&$80.60$ & $82.80$ \\
			CliNER \cite{boagcliner}&$79.50$&$81.20$ & $80.00$ \\
			truecasing CRFSuite \cite{fu2014improving}&$80.83$&$7 1.47$ & $75.86$ \\		
			\hline
			\bf(Our Approach) & & &  \\
			random-bidirectional LSTM-CRF &$00.00$ &$00.00$ & $78.13$ \\
			Word2Vec-bidirectional LSTM-CRF &$00.00$ &$00.00$ & $81.30$ \\
			Glove-bidirectional LSTM-CRF &$00.00$ &$00.00$ & $83.81$ \\
			\hline	
		\end{tabular}
		\caption{Performance comparison between the bidirectional LSTM CRF (bottom three lines) and state-of-the-art systems (top five lines) over the 2010 i2b2/VA concept extraction task.}
		\label{table3}
	\end{table*}

\section{Experiments}

\subsection{Datasets}
\label{sec:length}
The 2010 i2b2/VA Workshop on Natural Language Processing Challenges for Clinical Records presented three tasks, one among them is  concept extraction task focused on the extraction of medical concepts from patient  reports. A total of 394 training reports, 477 test reports, and 877 unannotated reports were de-identified and released to challenge participants with data use agreements~\cite{uzuner20112010}. However part of that data set is no longer being distributed due to Institutional Review Board (IRB) restrictions. Table~\ref{table2} summarizes the basic statistics of the training and test datasets used in our experiments. We split training dataset into a training and validation sets with approximately $70\%$ of sentences for training and the remaining for validation. 

\subsection{Evaluation Methodology}
\label{sec:length}
Our models have been blindly evaluated on unseen 2010 i2b2/VA  CE test data using the \textit{strict} evaluation metrics. With this evaluation, the predicted entities have to match the ground-truth entities exactly, both in boundary and class. To facilitate the replication of our experimental results, we have used a publicly-available library for the implementation (i.e., the Theano neural network toolkit \cite{bergstra2010theano}) and we publicly release our code\footnote{\hspace{-0.65cm}https://github.com/raghavchalapathy/Bidirectional-LSTM-CRF-for-Clinical-Concept-Extraction}. The experiments have been run over a range of values for the hyper-parameters, using the validation set for selection~\cite{bergstra2012random}. The hyper-parameters include the number of hidden-layer nodes, $H \in \{25, 50, 100\}$, the context window size, $s \in \{1, 3, 5\}$, and the embedding dimension, $d \in \{50, 100, 300, 500, 1000\}$. Two additional parameters, the learning and drop-out rates, were sampled from a uniform distribution in the range $[0.05, 0.1]$. To begin with, the embedding and initial weight matrices were all randomly initialized from the uniform distribution within range $[-1, 1]$ subsequently word embeddings with $d= 300$ derived from Word2Vec and Glove was utilized in the experiments. Early training stopping was set to $100$ epochs to mollify over-fitting, and the model that gave the best performance on the validation set was retained. The accuracy is reported in terms of micro-average F$_1$ score computed using the CoNLL score function~\cite{Nadeau:07}.

\subsection{Results and Analysis}

Table \ref{table3} shows the performance comparison between the employed bidirectional LSTM-CRF and state-of-the-art CE systems. As an overall note, the bidirectional LSTM-CRF have not reached the same accuracy as the top system, semi-supervised Markov HMM  \cite{de2011machine}. However, our approach has achieved the second-best score on 2010 i2b2/VA. These results seem interesting on the ground that the bidirectional LSTM-CRF provide CE without utilizing any manually-engineered features. Given that our system learn entirely from the data, it is also robust to any new concept or unseen words additions. In our current experimental setting about $20\%$ of tokens were either alpha-numeric or abbreviated strings whose Word2Vec or  Glove pretrained vector embeddings were not available. These special strings in text were randomly initialized with $d=300$ vector embeddings and input to bidirectional LSTM-CRF system. Subsequently the system was able to learn  meaningfull representations with remaining $80\%$ of pre-trained vector embeddings and produce comparable results to the state-of-the-art CE systems.

\label{sec:length}

\section*{Conclusion}

This paper has used the contemporary bidirectional LSTM-CRF, for clinical concept extraction. The most appealing feature of this sytem   is their ability to provide end-to-end recognition initialized with general purpose off-the-shelf word embeddings sparing effort from laborious feature construction. To the best of our knowledge, ours is the first paper to adopt bidirectional LSTM-CRF for concept extraction from clinical records. The experimental results over the  2010 i2b2/VA reference standard corpora look promising, with the bidirectional LSTM-CRF ranking closely to the state of the art. A potential way to further improve its performance would be to initialize its training with unsupervised word embeddings such as Word2Vec~\cite{Mikolov:13} and GloVe~\cite{Pennington:14} trained with domain specific resources such as Monitoring in Intensive Care (MIMIC) II copora. This approach has proved effective in many other domains and still dispenses with expert annotation effort; we plan this exploration for the near future. 

\bibliographystyle{acl}
\bibliography{coling2016}

\end{document}